\documentclass[journal,onecolumn]{IEEEtran}
\ifCLASSINFOpdf
\else
\fi
\usepackage{graphicx}
\usepackage{placeins}
\usepackage{mathtools}
\usepackage{listings}
\usepackage{threeparttable}
\graphicspath{ {./img/} }

\begin{document}
%
\title{Goods Transportation Problem Solving \\ via Routing Algorithm}
\author{Mikhail~Shchukin,~\IEEEmembership{Department of Computer Science,~University~of~Regina}\IEEEauthorrefmark{1}\\
        Aymen~Ben~Said,~\IEEEmembership{Department of Computer Science,~University~of~Regina}\IEEEauthorrefmark{2}\\
        André~Lobo~Teixeira,~\IEEEmembership{Department of Computer Science,~University~of~Regina}\IEEEauthorrefmark{3}
        \\
\IEEEauthorrefmark{1}vladimmi@uregina.ca, \IEEEauthorrefmark{2}aymenbensaid@uregina.ca, \IEEEauthorrefmark{3}teixeira@cs.uregina.ca}

\markboth{}%
{Shell \MakeLowercase{\textit{et al.}}: Design, Report, Computer Science}
%

\maketitle

\begin{abstract}
This paper outlines the ideas behind developing a graph-based heuristic-driven routing algorithm 
designed for a particular instance of a goods transportation problem with a single good type.
The proposed algorithm solves the optimization problem of satisfying the demand of goods on a given undirected
transportation graph with minimizing the estimated cost for each traversed segment of the delivery path. The operation
of the routing algorithm is discussed and overall evaluation of the proposed problem solving technique is given.
\end{abstract}


%
\IEEEpeerreviewmaketitle

\section{Introduction}
\IEEEPARstart{T}{he} transportation problem is one of the well-known and hot topics both in mathematics and economics.
It was first conceptualized by the French mathematician Gaspard Monge back in 1781 \cite{monge:monge}. Later in 1942, during the World War II, the Soviet mathematician and economist Leonid Kantorovich augmented and refreshed the mathematical definition of the problem from the linear programming point of view \cite{kanto:kanto}. Since then, many attempted to solve this problem using linear programming or graph-based algorithms. In this paper, a graph-based interpretation of the problem is used for reasons of better visualization of potential solutions (i.e. shipping paths) and the graph form being a comfortable environment to design the searching and routing algorithm. This paper strives to deal with this problem using graph theory approach and a particular decision-making algorithm in an attempt to acquire an acceptable solution, if it exists, or give partial solution otherwise.

From our point of view, the transportation problem can be represented by a graph with nodes of 3 types: \textit{store}, \textit{warehouse} and a \textit{joint}. The edges between nodes represent roads to traverse with associated \textit{distance} and \textit{time}. The stores and warehouses have a certain associated amount of \textit{demand} and \textit{supply} of goods respectively. Joints act solely as transfer points (major cities, towns, etc.). For simplicity, all goods are assumed to be of the \textit{single type}. The problem is finding the way to provide the stores with goods available in the warehouses, \textit{maximizing} the amount of delivered goods, while \textit{minimizing} the transportation costs. It is assumed that the shipping is being done by a \textit{single agent}, a truck with a certain \textit{maximum carry capacity}, that is traversing a certain path while loading and unloading goods where needed.

In this paper
, we provide an algorithm that takes a pre-generated transportation problem graph with the described characteristics and returns the following key elements of the proposed solution:
\begin{itemize}
	\item{Truck path with shipping and restocking segments}
	\item{Algorithm log with the following details:}
		\begin{itemize}
			\item{shipping and restocking segments explained}
			\item{routing decisions made for next segment (if applicable)}
			\item{decision cost (heuristic value)}
			\item{Remaining total amount of demand or supply at the termination of algorithm}
		\end{itemize}
	\item{Total runtime of the algorithm}
\end{itemize}

In the following sections we describe the testing environment for such algorithm, the mode of operation of such algorithm and the experimental results.

\section{Testing environment - graph generator}

First of all, in order to provide a capability of reproducing the experimental results of using the suggested algorithm, it is important to have a testing environment that can produce a set of various transportation problem instances (graphs) of a varying scale to serve as an input to the solving algorithm. The suggested testing environment attempts to pseudorandomly generate a transportation problem graph instance on each run. This allows to create a wide variety of sample cases that are applicable to the real-life instances of goods transportation problem.

The graph generator requires the following parameters to operate with:
\begin{itemize}
	\item{Total number of nodes to generate}
	\item{Maximum number of edges to spawn per node}
	\item{Total number of stores to generate}
	\item{Total number of warehouses to generate}
	\item{Types of goods to generate (in scope of this paper, this defaults to 1)}
	\item{Total supply and total demand per each good type}
\end{itemize}

Given these parameters, the graph generator proceeds to create a requested number of nodes. Each node is created as a joint by default. Then, the algorithm pseudorandomly decides which nodes are converted into stores and warehouses. To keep the graph realistic, each node receives pseudorandomly assigned coordinates of {X,Y}, where X,Y \(\in\) [0, MAX\_MAP\_SIZE], where MAX\_MAP\_SIZE corresponds to the scale of the square space where the transportation problem is defined, it defaults to 1000. Originally, it is assumed that the units of map scale are expressed in \textit{kilometers}. 

The generator then distributes goods to be supplied one by one between each warehouse until all supplies are allocated. Similarly, each store keeps incrementing the amount of demanded goods one by one until the demand is equally distributed among all stores. Equal distribution of supply and demand among stores and warehouses is assumed to avoid degenerate cases to happen (i.e. most goods are concentrated in a warehouse too close to the stores, one store demands much more goods than the others, etc.).

After the node generation is done, the algorithm spreads the required amount of edges. Each time an edge is to be created, there is a 50\% chance for that to happen. To avoid disconnected graphs, the algorithm forces at least one edge to be spread. The destination node is picked pseudorandomly among the rest of the generated nodes. When the edge is created, the {X,Y} coordinates of the source and destination are used to obtain the \textit{distance between nodes}. Then, an \textit{average assumed velocity} is picked pseudorandomly within the range of [40,100]. These bounds are assuming that the velocity is expressed in \textit{kilometers per hour}, where lower bound corresponds to speed limits of bigger cities and upper bound resembles the highways. Based on the average assumed velocity of the edge, the time is then calculated as 
\begin{equation*}
time\_in\_minutes = (distance\_between\_nodes / assumed\_average\_velocity) \times 60
\end{equation*}
with floating point of it truncated. The time and distance are the two key parameters that affect the later decision-making, so they are kept in the graph as the background information.

\begin{figure}[h]
	\centering
  	\includegraphics[width=0.5\textwidth]{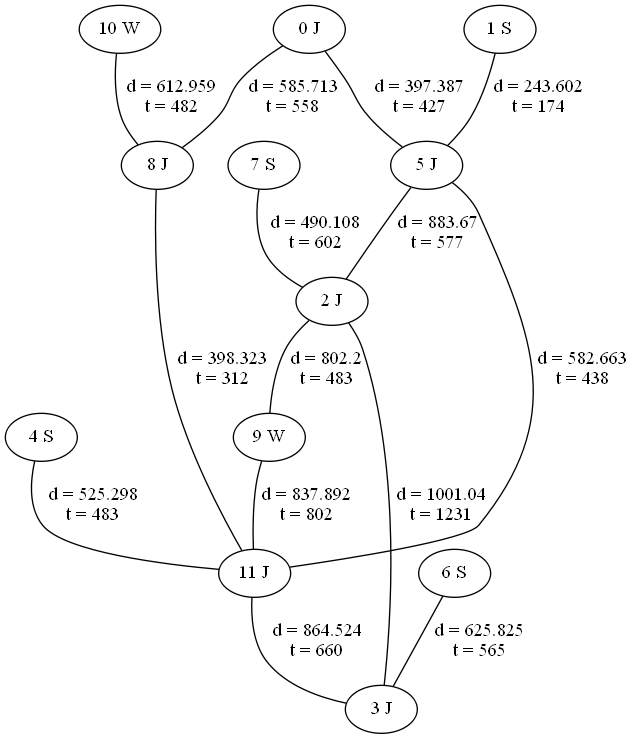}
  	\caption{An example of transportation graph produced by the graph generator}	
  	\label{fig:img1}
\end{figure}

When edges between nodes are created, the graph generator produces a simple text file, that has the problem-related information associated with nodes and edges. Notably, the syntax for such text file is specifically designed to follow the syntax of GraphViz, the graph visualization tool that can produce and display the graph, based on the plain text graph description with all embedded information as a picture \cite{web:web}. A particular example of such visualization can be seen on Fig. \ref{fig:img1}.

Evidently, the graph generator provides a fairly connected graph that represents a particular instance of a transportation problem. Each node shows a tuple of \textit{\{id,type\}}. Node 0 is the first generated node and the total number of transportation nodes in this particular example is 12. To generate this example, the graph generator is initialized with 2 edges to spread from each node, 2 warehouses and 4 stores to be placed. The total goods demand is set to 90 and the total supply of goods is set to 100, which is spread equally across stores and warehouses respectively. Note, that in order to reduce the visible complexity of the picture, the visual part does not contain all the problem parameters, such as supply or demand, straight on it. Nevertheless, all crucial problem parameters are still stored in the source file structure, as can be seen on Listing \ref{lst1}.

\lstset{basicstyle=\ttfamily}
\begin{lstlisting}[language=C,breaklines=true,caption={The underlying textual form of the graph},captionpos=b,showstringspaces=false,label=lst1]
graph G {
"0" [label="0 J", type=0, supply=0, demand=0]
"0"--"5"[label=" d = 397.387\n t = 427", distance=397.387, time=427]
"0"--"8"[label=" d = 585.713\n t = 558", distance=585.713, time=558]
"1" [label="1 S", type=1, supply=0, demand=22]
"1"--"5"[label=" d = 243.602\n t = 174", distance=243.602, time=174]
"2" [label="2 J", type=0, supply=0, demand=0]
"2"--"9"[label=" d = 802.2\n t = 483", distance=802.2, time=483]
"3" [label="3 J", type=0, supply=0, demand=0]
"3"--"2"[label=" d = 1001.04\n t = 1231", distance=1001.04, time=1231]
"4" [label="4 S", type=1, supply=0, demand=22]
"4"--"11"[label=" d = 525.298\n t = 483", distance=525.298, time=483]
"5" [label="5 J", type=0, supply=0, demand=0]
"5"--"2"[label=" d = 883.67\n t = 577", distance=883.67, time=577]
"5"--"11"[label=" d = 582.663\n t = 438", distance=582.663, time=438]
"6" [label="6 S", type=1, supply=0, demand=23]
"6"--"3"[label=" d = 625.825\n t = 565", distance=625.825, time=565]
"7" [label="7 S", type=1, supply=0, demand=23]
"7"--"2"[label=" d = 490.108\n t = 602", distance=490.108, time=602]
"8" [label="8 J", type=0, supply=0, demand=0]
"8"--"11"[label=" d = 398.323\n t = 312", distance=398.323, time=312]
"9" [label="9 W", type=2, supply=50, demand=0]
"9"--"11"[label=" d = 837.892\n t = 802", distance=837.892, time=802]
"10" [label="10 W", type=2, supply=50, demand=0]
"10"--"8"[label=" d = 612.959\n t = 482", distance=612.959, time=482]
"11" [label="11 J", type=0, supply=0, demand=0]
"11"--"3"[label=" d = 864.524\n t = 660", distance=864.524, time=660]
}
\end{lstlisting}

This testing environment combines the commodity of reproducable results, configurable problem scaling and the power of visualization through external tools. In the following sections, we refer to Fig. \ref{fig:img1} as a particular case of using this testing environment.

\section{The Routing Algorithm}

Given an instance of the goods transportation graph, the routing algorithm requires the following parameters to attempt at solving the problem:
\begin{itemize}
	\item{Truck initial position (defaults to node 0, which is the first generated node)}
	\item{Truck initial load of supplies (defaults to 0, we assume here that the truck always starts empty)}
	\item{Truck maximum carry capacity (MAX\_CAPACITY as referred below)}
	\item{Resupply threshold \textbf{T} (level of current supply below which a truck restocks at a warehouse, not proceeding to stores)}
\end{itemize}

With these parameters set, the truck attempts to follow the least expensive path, which is recorded as a solution. The heuristic cost that is used in determination of the least expensive edge to traverse is defined as follows: 

\begin{equation*}
edge\_cost = time + distance
\end{equation*}

Accordingly, at the beginning of each segment of the traversed path the truck driver decision-making works as follows:
\begin{itemize}
    \item{If the position of the truck is at a warehouse node with a capacity \textless  \ \textbf{T} $ \times $ MAX\_CAPACITY : it goes to the nearest warehouse which is the current node and restocks right there.}
    \item{If the position of the truck is at a joint node with a capacity \textless \ \textbf{T} $ \times $ MAX\_CAPACITY: it goes to the nearest warehouse to restock with goods.}
    \item{If the position of the truck  is at a store node with a capacity \textless  \ \textbf{T} $ \times $ MAX\_CAPACITY: it goes to the nearest warehouse to restock with goods.}
   \item{If the position of the truck is at a warehouse node with a capacity \textgreater= \textbf{T} $ \times $ MAX\_CAPACITY: it goes to the nearest store to drop goods.}
   \item{If the position of the truck is at a joint node with a capacity \textgreater= \textbf{T} $ \times $ MAX\_CAPACITY: it goes to the nearest store to drop goods.}
   \item{If the position of the truck is at a store node with a capacity \textgreater= \textbf{T} $ \times $ MAX\_CAPACITY: it goes and drops goods at the current node.}  
\end{itemize}

Note that for the sake of simplicity, we assume \textbf{T} = 0.5 for the particular instance of the graph shown in Fig. \ref{fig:img1}.

The algorithm treats each node according to their type with 2 parameters: supply and demand. For warehouse nodes, the demand is always 0; for stores, the supply is always 0; for joints, both supply and demand are 0. 

The truck keeps going back and forth between the stores and the warehouses until either all warehouses deplete their supply (which indicates an \textit{overconstrained problem} with only a partial solution available) or there remains no store still waiting for goods (which indicates a \textit{successful termination} of the algorithm with all goods shipped to the stores).

Thus, if the graph shown in Listing \ref{fig:img1} is given for this algorithm with truck maximum capacity set to 20, the result shown on Listing \ref{lst2} in Appendix \ref{app1} is produced and saved as an output plain text file.

As seen from Listing \ref{lst2}, the algorithm provides a detailed breakdown of the logic of its operation. The design of the algorithm includes a self-explanatory solver, this algorithm gives its user complete information of the state of the program at each decision-making stage. 
At the end of the exhaustive report, the algorithm provides the path traversed by the truck as well as the final state of the operation. In this particular instance, since the total demand is 90 and is  less than total supply of 100, the solution is complete and as seen in the resulting file, the total demand and supply are both zero with 10 units of cargo remaining in the truck.

\begin{figure}[h]
	\centering
  	\includegraphics[width=0.5\textwidth]{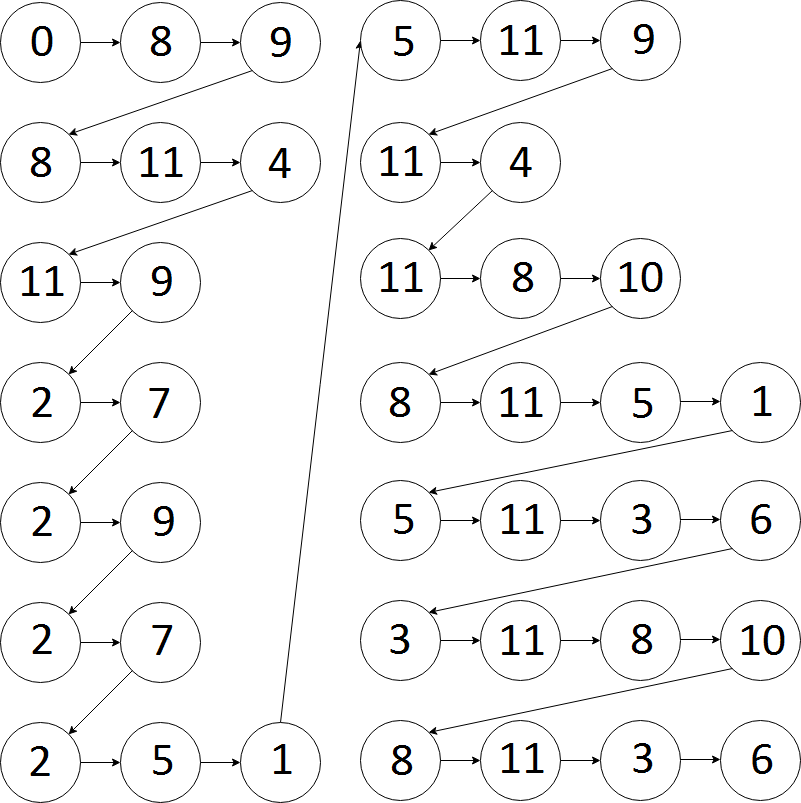}
  	\caption{The transportation solution path constructed by amalgamating supply and restock path segments}	
  	\label{fig:img4}
\end{figure}

Fig. \ref{fig:img4} demonstrates the solution path that the described algorithm suggests for a given instance of the problem acquired through the testing environment that produced the base graph with transportation problem parameters.

\section{Conclusion}
After observing the fundamenal operation principles of the proposed algorithm, it is important to note that such algorithm is capable of functioning with a wide spectrum of transportation problem instances. Under experimental conditions, the given algorithm launched on the demonstrated graph  with variable truck capacity produces economically useful and mathematically reasonable results. As seen from Table \ref{tab:tab1}, with increasing the truck capacity, the number of produced path segments and the cost of the overall path decreases.
 
\begin{table}[h]
\centering
\caption{Routing algorithm experimental results} 
\label{tab:tab1}
	\begin{tabular}{ | l | l | l |}
 		\hline
 		Truck Capacity & Number of path segments & Overall path cost \\ \hline
 		10 & 22 segments & 68360.5  \\ \hline
		15 & 16 segments & 48142.7 \\ \hline
		20 & 14 segments & 42781.1  \\ \hline
		22 & 13 segments & 40764.5 \\ \hline
		23 & 9 segments & 26577.2 \\ \hline
	\end{tabular}
\end{table}

However, it is notable that the change is not quite proportional. As seen on Fig. \ref{fig:img2} and Fig. \ref{fig:img3} below, changing the truck capacity produces a slope-like response in the cost reduction and in the number of path segments produced.


\begin{figure}[h]
	\centering
  	\includegraphics[width=0.6\textwidth]{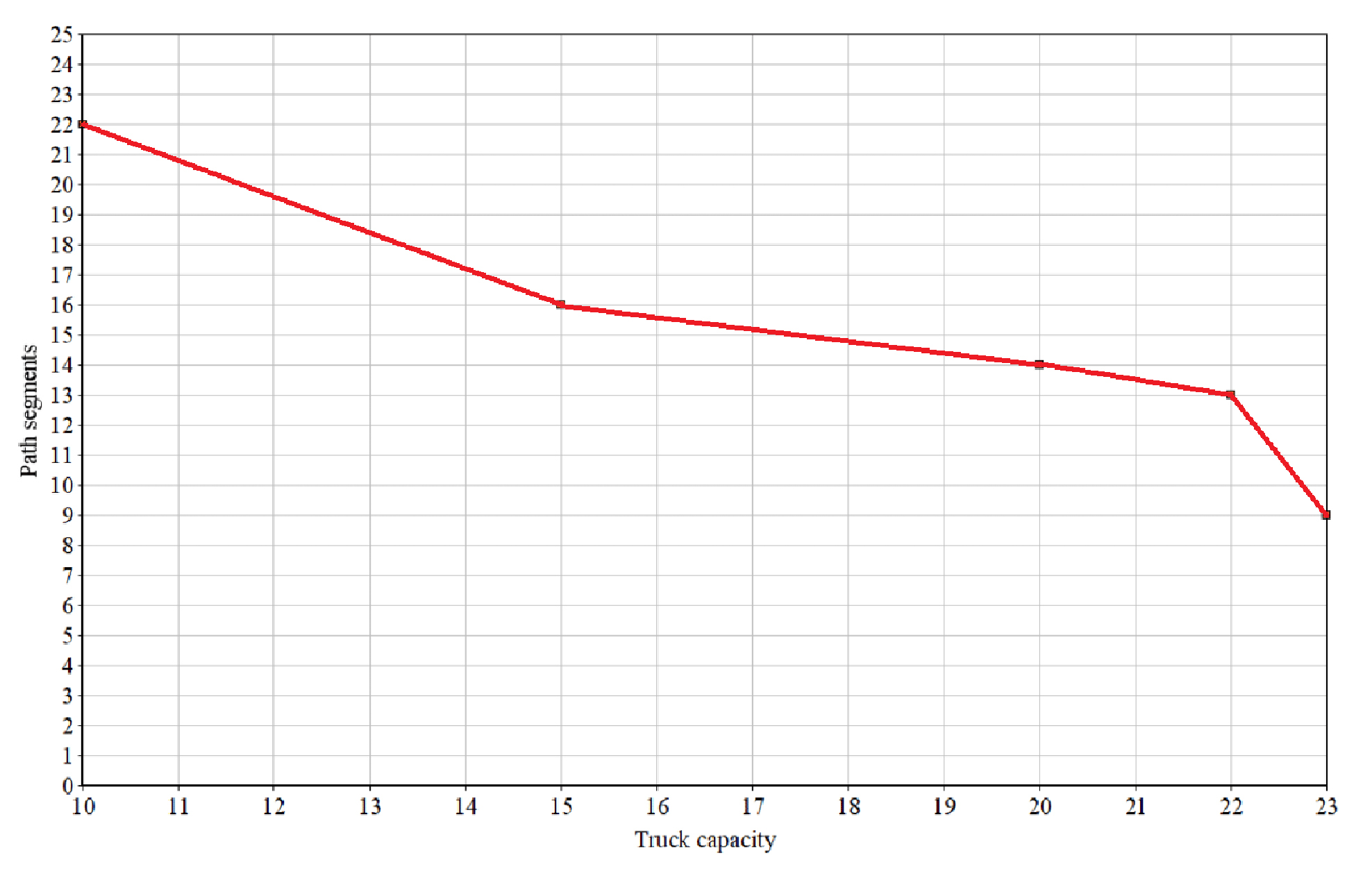}
  	\caption{Truck capacity increase reduces the number of shipping path segments}	
  	\label{fig:img2}
\end{figure}

\begin{figure}[h]
	\centering
  	\includegraphics[width=0.6\textwidth]{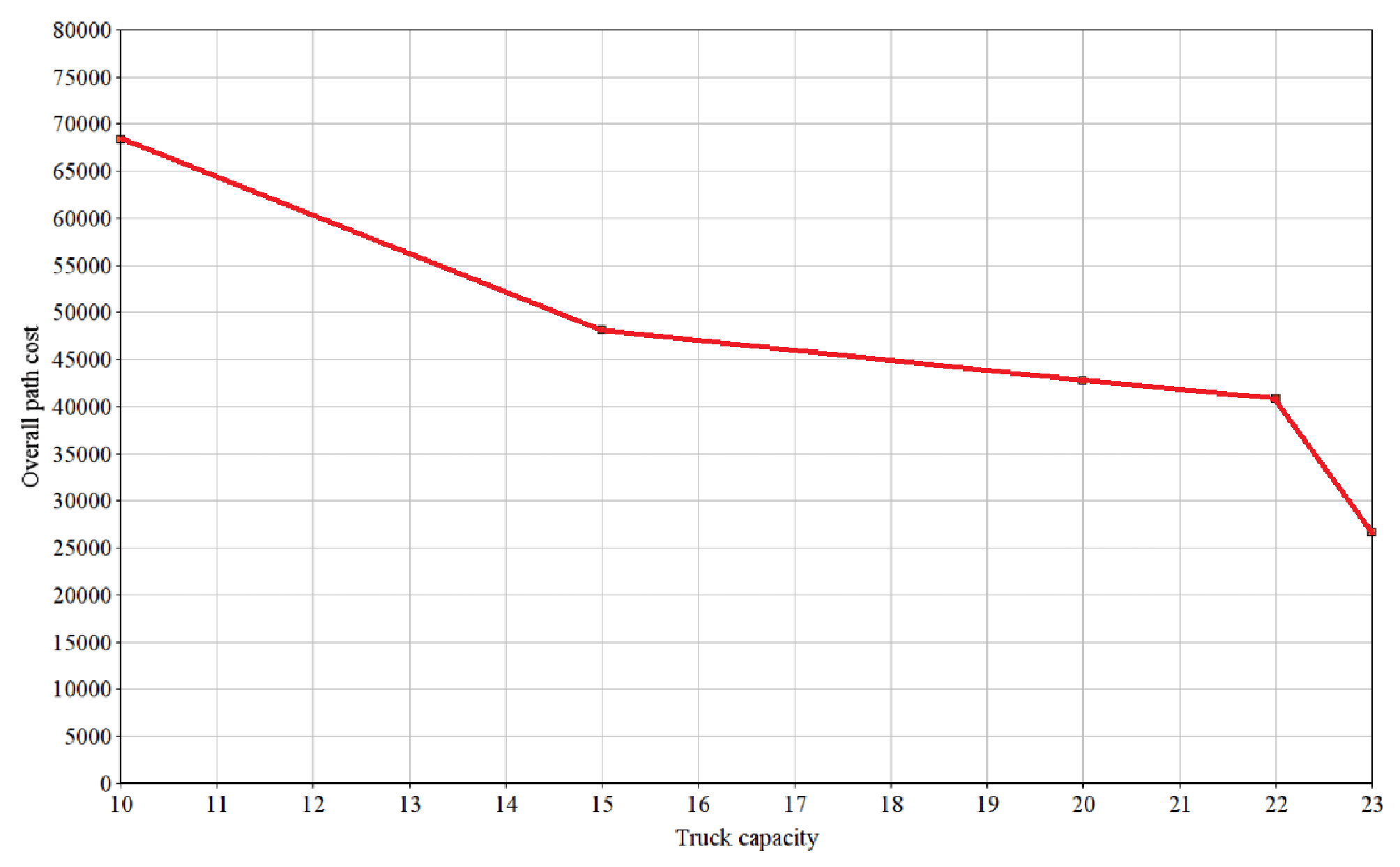}
  	\caption{Truck capacity increase gradually reduces the overall path cost}	
  	\label{fig:img3}
\end{figure}

Given such statistics, it is possible to estimate how much each added unit of capacity corresponds to the decrease in the overall solution cost for this particular instance of the problem. This directly translates into the economic reality of any business involving transportation of some items between certain locations on the map. Such algorithm is thus capable to provide not only a solution to the given problem with set parameters, but also \textit{an ability to estimate the potential alternatives to a given solution with different parameters}. 

What if one had to pick a bigger truck? How could extra expenses on deploying another truck be covered by saving on the number of trips between the map locations? What is the most efficient size of the truck, that can solve a given transportation problem on a given map? All these questions can eventually be answered by running the routing algorithm provided here several times with tweaked parameters. Basically, by solving the given problem with different parameters and combining the resulting data together, this algorithm can be used as a \textit{cost estimation tool}.

There are, however, major improvements potentially available to this algorithm. For example, the cost estimation function could involve more real-life factors affecting any transportation, such as: truck fuel consumption, desirability of paths (major city roads might not allow truck traffic coming through at certain times), required stops (drivers taking breaks and changing shifts) and more. Another improvement to deploy could be allowing multiple types of cargo, which considerably increases the complexity of decision-making and solving the transportation problem.

In the described algorithm, the decision-making heuristic of combined time and distance is only applied to selecting next place to go, so the overall cost is minimized per each segment of the solution path. However, sometimes the transportation problem can involve prioritizing time over cost, cost over time or other decision factor. Thus, wisely balancing the decision-making in the described routing algorithm along with dealing with various shipment types becomes the major design difficulty to overcome.

\ifCLASSOPTIONcaptionsoff
  \newpage
\fi

\appendices

\section{Output listing}
\label{app1}
\lstset{basicstyle=\ttfamily}
\begin{lstlisting}[language=C,breaklines=true,caption={The output of the routing algorithm solving the problem},captionpos=b,showstringspaces=false,label=lst2]
----------- Iteration 1: Truck at node 0
Current STATUS:
Total Demand: 90/90  Demand list: [22, 22, 23, 23]
Total Supply: 100/100  Supply list: [50, 50]
Truck Current Load: 0/20


----- A) Decision: GOTO A WAREHOUSE, as the load 0 is bellow the threshold 10.0:

----- B) Searching the warehouse with the least path cost >>>
Supplies list:[50, 50]
--- current W id: 9
--- path to current W: [0, 5, 11, 9]
--- path cost: 3484.942

--- current W id: 10
--- path to current W: [0, 8, 10]
--- path cost: 2238.6719999999996

******Final Decision:
GOTO warehouse 10
GET 20 supplies from this warehouse


----------- Iteration 2: Truck at node 10
Current STATUS:
Total Demand: 90/90  Demand list: [22, 22, 23, 23]
Total Supply: 80/100  Supply list: [50, 30]
Truck Current Load: 20/20


----- A) Decision: GOTO A STORE, as the load 20 is above the threshold 10.0:

----- B) Searching the store with least path cost >>>
Demand list:[22, 22, 23, 23]
--- current S id: 1
--- path to current S: [10, 8, 11, 5, 1]
--- path cost: 3243.5469999999996

--- current S id: 4
--- path to current S: [10, 8, 11, 4]
--- path cost: 2813.58

--- current S id: 6
--- path to current S: [10, 8, 11, 3, 6]
--- path cost: 4520.630999999999

--- current S id: 7
--- path to current S: [10, 8, 11, 5, 2, 7]
--- path cost: 5378.723

******Final Decision:
GOTO store 4
SUPPLY the store with the current truck load of 20


----------- Iteration 3: Truck at node 4
Current STATUS:
Total Demand: 70/90  Demand list: [22, 2, 23, 23]
Total Supply: 80/100  Supply list: [50, 30]
Truck Current Load: 0/20


----- A) Decision: GOTO A WAREHOUSE, as the load 0 is bellow the threshold 10.0:

----- B) Searching the warehouse with the least path cost >>>
Supplies list:[50, 30]
--- current W id: 9
--- path to current W: [4, 11, 9]
--- path cost: 2648.19

--- current W id: 10
--- path to current W: [4, 11, 8, 10]
--- path cost: 2813.58

******Final Decision:
GOTO warehouse 9
GET 20 supplies from this warehouse


----------- Iteration 4: Truck at node 9
Current STATUS:
Total Demand: 70/90  Demand list: [22, 2, 23, 23]
Total Supply: 60/100  Supply list: [30, 30]
Truck Current Load: 20/20


----- A) Decision: GOTO A STORE, as the load 20 is above the threshold 10.0:

----- B) Searching the store with least path cost >>>
Demand list:[22, 2, 23, 23]
--- current S id: 1
--- path to current S: [9, 11, 5, 1]
--- path cost: 3078.157

--- current S id: 4
--- path to current S: [9, 11, 4]
--- path cost: 2648.19

--- current S id: 6
--- path to current S: [9, 11, 3, 6]
--- path cost: 4355.241

--- current S id: 7
--- path to current S: [9, 2, 7]
--- path cost: 2377.308

******Final Decision:
GOTO store 7
SUPPLY the store with the current truck load of 20


----------- Iteration 5: Truck at node 7
Current STATUS:
Total Demand: 50/90  Demand list: [22, 2, 23, 3]
Total Supply: 60/100  Supply list: [30, 30]
Truck Current Load: 0/20


----- A) Decision: GOTO A WAREHOUSE, as the load 0 is bellow the threshold 10.0:

----- B) Searching the warehouse with the least path cost >>>
Supplies list:[30, 30]
--- current W id: 9
--- path to current W: [7, 2, 9]
--- path cost: 2377.308

--- current W id: 10
--- path to current W: [7, 2, 5, 11, 8, 10]
--- path cost: 5378.723

******Final Decision:
GOTO warehouse 9
GET 20 supplies from this warehouse


----------- Iteration 6: Truck at node 9
Current STATUS:
Total Demand: 50/90  Demand list: [22, 2, 23, 3]
Total Supply: 40/100  Supply list: [10, 30]
Truck Current Load: 20/20


----- A) Decision: GOTO A STORE, as the load 20 is above the threshold 10.0:

----- B) Searching the store with least path cost >>>
Demand list:[22, 2, 23, 3]
--- current S id: 1
--- path to current S: [9, 11, 5, 1]
--- path cost: 3078.157

--- current S id: 4
--- path to current S: [9, 11, 4]
--- path cost: 2648.19

--- current S id: 6
--- path to current S: [9, 11, 3, 6]
--- path cost: 4355.241

--- current S id: 7
--- path to current S: [9, 2, 7]
--- path cost: 2377.308

******Final Decision:
GOTO store 7
SUPPLY the entire demand from this store


----------- Iteration 7: Truck at node 7
Current STATUS:
Total Demand: 47/90  Demand list: [22, 2, 23, 0]
Total Supply: 40/100  Supply list: [10, 30]
Truck Current Load: 17/20


----- A) Decision: GOTO A STORE, as the load 17 is above the threshold 10.0:

----- B) Searching the store with least path cost >>>
Demand list:[22, 2, 23, 0]
--- current S id: 1
--- path to current S: [7, 2, 5, 1]
--- path cost: 2970.38

--- current S id: 4
--- path to current S: [7, 2, 5, 11, 4]
--- path cost: 4581.7390000000005

--- current S id: 6
--- path to current S: [7, 2, 3, 6]
--- path cost: 4514.973

******Final Decision:
GOTO store 1
SUPPLY the store with the current truck load of 17


----------- Iteration 8: Truck at node 1
Current STATUS:
Total Demand: 30/90  Demand list: [5, 2, 23, 0]
Total Supply: 40/100  Supply list: [10, 30]
Truck Current Load: 0/20


----- A) Decision: GOTO A WAREHOUSE, as the load 0 is bellow the threshold 10.0:

----- B) Searching the warehouse with the least path cost >>>
Supplies list:[10, 30]
--- current W id: 9
--- path to current W: [1, 5, 11, 9]
--- path cost: 3078.157

--- current W id: 10
--- path to current W: [1, 5, 11, 8, 10]
--- path cost: 3243.5469999999996

******Final Decision:
GOTO warehouse 9
GET all supplies from this warehouse


----------- Iteration 9: Truck at node 9
Current STATUS:
Total Demand: 30/90  Demand list: [5, 2, 23, 0]
Total Supply: 30/100  Supply list: [0, 30]
Truck Current Load: 10/20


----- A) Decision: GOTO A STORE, as the load 10 is above the threshold 10.0:

----- B) Searching the store with least path cost >>>
Demand list:[5, 2, 23, 0]
--- current S id: 1
--- path to current S: [9, 11, 5, 1]
--- path cost: 3078.157

--- current S id: 4
--- path to current S: [9, 11, 4]
--- path cost: 2648.19

--- current S id: 6
--- path to current S: [9, 11, 3, 6]
--- path cost: 4355.241

******Final Decision:
GOTO store 4
SUPPLY the entire demand from this store


----------- Iteration 10: Truck at node 4
Current STATUS:
Total Demand: 28/90  Demand list: [5, 0, 23, 0]
Total Supply: 30/100  Supply list: [0, 30]
Truck Current Load: 8/20


----- A) Decision: GOTO A WAREHOUSE, as the load 8 is bellow the threshold 10.0:

----- B) Searching the warehouse with the least path cost >>>
Supplies list:[0, 30]
--- current W id: 10
--- path to current W: [4, 11, 8, 10]
--- path cost: 2813.58

******Final Decision:
GOTO warehouse 10
GET 12 supplies from this warehouse


----------- Iteration 11: Truck at node 10
Current STATUS:
Total Demand: 28/90  Demand list: [5, 0, 23, 0]
Total Supply: 18/100  Supply list: [0, 18]
Truck Current Load: 20/20


----- A) Decision: GOTO A STORE, as the load 20 is above the threshold 10.0:

----- B) Searching the store with least path cost >>>
Demand list:[5, 0, 23, 0]
--- current S id: 1
--- path to current S: [10, 8, 11, 5, 1]
--- path cost: 3243.5469999999996

--- current S id: 6
--- path to current S: [10, 8, 11, 3, 6]
--- path cost: 4520.630999999999

******Final Decision:
GOTO store 1
SUPPLY the entire demand from this store


----------- Iteration 12: Truck at node 1
Current STATUS:
Total Demand: 23/90  Demand list: [0, 0, 23, 0]
Total Supply: 18/100  Supply list: [0, 18]
Truck Current Load: 15/20


----- A) Decision: GOTO A STORE, as the load 15 is above the threshold 10.0:

----- B) Searching the store with least path cost >>>
Demand list:[0, 0, 23, 0]
--- current S id: 6
--- path to current S: [1, 5, 11, 3, 6]
--- path cost: 4153.614

******Final Decision:
GOTO store 6
SUPPLY the store with the current truck load of 15


----------- Iteration 13: Truck at node 6
Current STATUS:
Total Demand: 8/90  Demand list: [0, 0, 8, 0]
Total Supply: 18/100  Supply list: [0, 18]
Truck Current Load: 0/20


----- A) Decision: GOTO A WAREHOUSE, as the load 0 is bellow the threshold 10.0:

----- B) Searching the warehouse with the least path cost >>>
Supplies list:[0, 18]
--- current W id: 10
--- path to current W: [6, 3, 11, 8, 10]
--- path cost: 4520.630999999999

******Final Decision:
GOTO warehouse 10
GET all supplies from this warehouse


----------- Iteration 14: Truck at node 10
Current STATUS:
Total Demand: 8/90  Demand list: [0, 0, 8, 0]
Total Supply: 0/100  Supply list: [0, 0]
Truck Current Load: 18/20


----- A) Decision: GOTO A STORE, as the load 18 is above the threshold 10.0:

----- B) Searching the store with least path cost >>>
Demand list:[0, 0, 8, 0]
--- current S id: 6
--- path to current S: [10, 8, 11, 3, 6]
--- path cost: 4520.630999999999

******Final Decision:
GOTO store 6
SUPPLY the entire demand from this store



------------ Final State of global demand and supply:
Remainging demand: 0
Remainging supplies: 0


------------ Final State of global demand and supply

Total time of the experiment:0.00540311899385415


Truck started at node (0) with initial load of (0), and capacity of (20)

The algorithm used a threshold factor of (0.5)

The imported graph file name is 'graph'

The overall cost of the truck's path is: 42781.09599999998
The list of paths is: 
[0, 8, 10]
[10, 8, 11, 4]
[4, 11, 9]
[9, 2, 7]
[7, 2, 9]
[9, 2, 7]
[7, 2, 5, 1]
[1, 5, 11, 9]
[9, 11, 4]
[4, 11, 8, 10]
[10, 8, 11, 5, 1]
[1, 5, 11, 3, 6]
[6, 3, 11, 8, 10]
[10, 8, 11, 3, 6]
\end{lstlisting}




\begin{thebibliography}{1}

\bibitem{monge:monge}
G. Monge. \emph{Mémoire sur la théorie des déblais et des remblais. Histoire de l’Académie Royale des Sciences de Paris, avec les Mémoires de Mathématique et de Physique pour la même année}. , pages 666–704, 1781.

\bibitem{kanto:kanto}
L. Kantorovich. \emph{On the translocation of masses}. C.R. (Doklady) Acad. Sci. URSS (N.S.), 37:199–201, 1942.

\bibitem{web:web}
Graphviz.org. \emph{Graphviz - Graph Visualization Software}. 2019. [online] Available at: http://www.graphviz.org/ [Accessed 24 Mar. 2019].

\end{thebibliography}
%



\end{document}